\documentclass[final]{cvpr}

\usepackage{times}
\usepackage{epsfig}
\usepackage{graphicx}
\usepackage{amsmath}
\usepackage{amssymb}
\usepackage{tabu}
\usepackage{multirow}
\usepackage{xcolor}
\usepackage{colortbl}
\usepackage{subcaption}

\definecolor{light-gray}{gray}{0.75}

\usepackage[pagebackref=true,breaklinks=true,colorlinks,bookmarks=false]{hyperref}

\def\cvprPaperID{1846} 
\def\confYear{CVPR 2021}

\begin{document}

\title{Self-supervised Video Representation Learning by Context and Motion Decoupling}
\author{Lianghua Huang, Yu Liu, Bin Wang, Pan Pan, Yinghui Xu, Rong Jin\\
Machine Intelligence Technology Lab, Alibaba Group\\
{\tt\small xuangen.hlh,ly103369,ganfu.wb,panpan.pp,renji.xyh,jinrong.jr@alibaba-inc.com}
}

\maketitle

\begin{abstract}
   A key challenge in self-supervised video representation learning is how to effectively capture motion information besides context bias.
   While most existing works implicitly achieve this with video-specific pretext tasks (e.g., predicting clip orders, time arrows, and paces), we develop a method that explicitly decouples motion supervision from context bias through a carefully designed pretext task.
   Specifically, we take the key frames and motion vectors in compressed videos (e.g., in H.264 format) as the supervision sources for context and motion, respectively, which can be efficiently extracted at over 500 fps on CPU.
   Then we design two pretext tasks that are jointly optimized:
   a \textbf{context matching} task where a pairwise contrastive loss is cast between video clip and key frame features;
   and a \textbf{motion prediction} task where clip features, passed through an encoder-decoder network, are used to estimate motion features in a near future.
   These two tasks use a shared video backbone and separate MLP heads.
   Experiments show that our approach improves the quality of the learned video representation over previous works, where we obtain absolute gains of $16.0\%$ and $11.1\%$ in video retrieval recall on UCF101 and HMDB51, respectively.
   Moreover, we find the motion prediction to be a strong regularization for video networks, where using it as an auxiliary task improves the accuracy of action recognition with a margin of $7.4\%\sim 13.8\%$.
\end{abstract}

\section{Introduction}

\begin{figure}[t]
\begin{center}
\includegraphics[width=0.95\linewidth]{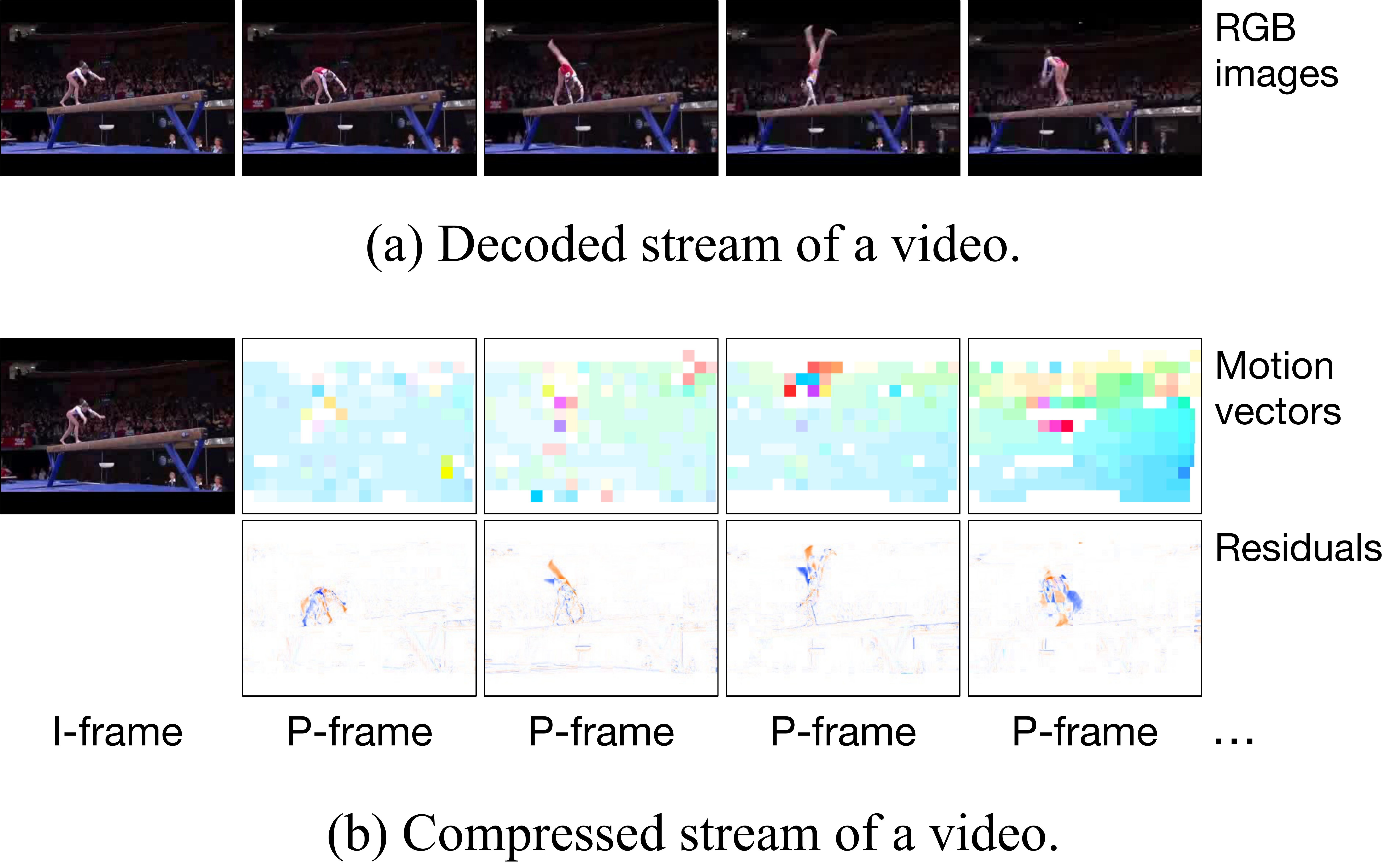}
\end{center}
   \caption{
      (a) and (b) show the decoded and compressed streams of a sample video, respectively.
      We notice that the \textbf{context} and \textbf{motion} information are roughly decoupled in I-frames and motion vectors of the compressed stream.
      We exploit these modalities as the supervision sources for self-supervised video representation learning.
      }
\label{fig:overview_compressed}
\end{figure}

Self-supervised representation learning from unlabeled videos has recently received considerable attention~\cite{memorydpc2020,vtdl2020}.
Compared with static images, the redundancy, temporal consistency, and multi-modality of videos potentially provide richer sources of ``supervision''.
Various methods have been proposed in this field that learn video representation by
designing video-specific pretext tasks~\cite{shuffle_learn2016,vcop2019,pacepred2020},
adapting contrastive learning to videos~\cite{dpc2019,vtdl2020,cvrl2020,memorydpc2020},
cross-modal learning~\cite{atvs2018,videobert2019,gdt2020,milnce2020},
and contrastive clustering~\cite{xdc2019}.

\begin{figure*}[t]
\begin{center}
\includegraphics[width=0.96\linewidth]{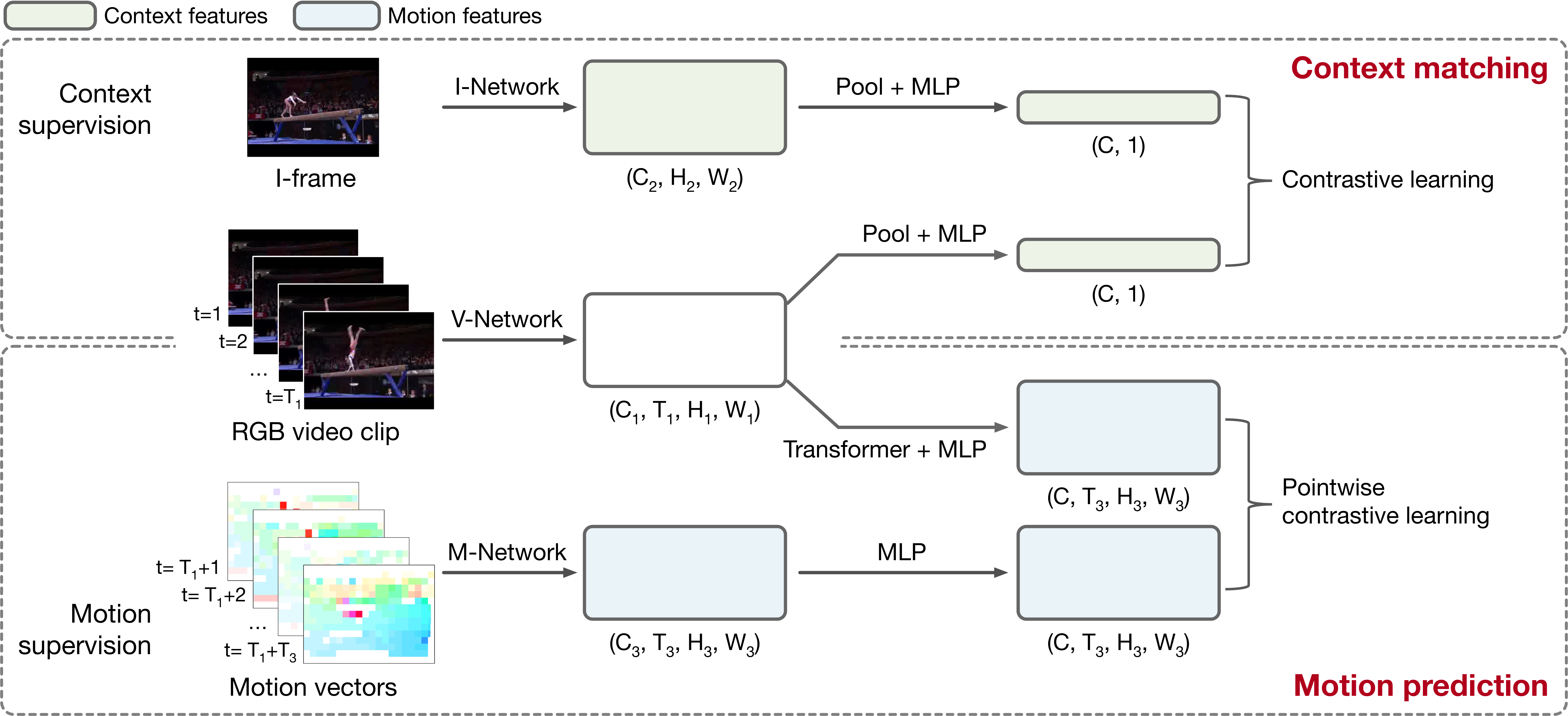}
\end{center}
   \caption{
      An overview of our framework.
      We decouple the context and motion supervision in two separate pretext tasks: context matching and motion prediction.
      The context matching task takes the relatively static I-frames in the compressed video as the source of supervision,
      and casts a contrastive loss between global features of I-frames and video clips.
      The motion prediction task takes the motion vector maps extracted from \textit{future frames} of the compressed video as the source of supervision,
      and compares the predicted and ``groundtruth'' motion features in a pointwise way using the contrastive loss.
      }
\label{fig:overview}
\end{figure*}

This paper focuses on visual-only video representation learning, and we distinguish two orthogonal but complementary aspects of video representation: \textbf{context} and \textbf{motion}.
Context depicts coarse-grained and relatively static environments, while motion represents dynamic and fine-grained movements or actions.
Context information alone can be used to classify certain actions (\eg, \textit{swimming} is most likely to take place at \textit{swimming pool}), but it also leads to background bias~\cite{twostream2014,diving48_2018}.
For actions that heavily depend on movement patterns (\eg, \textit{breaststroke} and \textit{frontcrawl}), motion information must be introduced.
We aim to design a self-supervised video representation learning method that jointly learns these two complementary information.
Our idea is to design a multi-task framework, where context and motion representation learning is decoupled in pretext tasks.
One problem here is the source of supervision.
Considering the scalability of our framework on larger datasets, we endeavor to avoid the use of computationally expensive features such as optical flow~\cite{tvl1_2007,highacc2004} and dense trajectories~\cite{densetraj2011}.

We notice that video in compressed format (such as H.264 and MPEG-4) roughly decouples the context and motion information in its I-frames and motion vectors.
As shown in Figure~\ref{fig:overview_compressed}, a compressed video stores only a few key frames (\ie, I-frames) completely, and it reconstructs other frames based on motion vectors (\ie, pixel offsets) and residual errors from the key frames.
I-frames can represent relatively static and coarse-grained context information, while motion vectors depict dynamic and fine-grained movements.
Moreover, both modalities can be efficiently extracted at over 500 fps on CPU~\cite{coviar2018}.

Inspired by this, we present a self-supervised video representation learning method where two decoupled pretext tasks are jointly optimized: \textbf{context matching} and \textbf{motion prediction}.
Figure~\ref{fig:overview} shows an overview of our framework.
The \textit{context matching} task aims to give the video network a rough grasp of the environment in which actions take place.
It casts a noise contrastive estimation (NCE) loss~\cite{nce2010,infonce2018} between global features of video clips and I-frames, where clips and I-frames from the same videos are pulled together, while those from different videos are pushed away.
The \textit{motion prediction} task requires the model to predict pointwise motion dynamics \textit{in a near future} based on visual information of the current clip.
The assumption is that, in order to predict future motion, the video network needs to extract \textit{low-level} movements from visual data and reorganize them into \textit{high-level} trajectories.
In this way, the learned representation should contain semantic and long-term motion information, helpful for downstream tasks.
In our framework, instead of directly estimating the values of motion vectors, we use pointwise contrastive learning to compare predicted and real motion features at every spatial and temporal location $(x, y, t)$.
We find that this leads to more stable pretraining and better transferring performance.

We conduct extensive experiments on three network architectures, three datasets, and two downstream tasks (\ie, action recognition and video retrieval) to assess the quality of our learned video representation.
We achieve state-of-the-art performance on all these experiments.
For example, we achieve R@1 video retrieval scores of 41.7\% and 16.8\% respectively on UCF101~\cite{ucf101} and HMDB51~\cite{hmdb51} datasets, obtaining 16.0\% and 11.1\% absolute gains compared to existing works.
We also validate several modeling options in our ablation studies,
where we find that the \textit{motion prediction} can serve as a strong regularization for video networks, and using it as an auxiliary task clearly improves the performance of supervised action recognition.

We summarize our contributions in the following:
\begin{itemize}
\setlength\itemsep{0em}
   \item Unlike existing works where the source of supervision usually comes from the decoded raw video frames, we present a self-supervised video representation learning method that explicitly decouples the context and motion supervision in the pretext task.
   \item We present a \textit{context matching} task for learning coarse-grained and relatively static context representation, and a \textit{motion prediction} task for learning fine-grained and high-level motion representation.
   \item To the best of our knowledge, we present the first approach that exploits the modalities in compressed videos as the efficient supervision sources for visual representation learning.
   \item We achieve significant improvements over existing works on downstream tasks of action recognition and video retrieval. Extensive ablation studies also validate the effectiveness of our several modeling options.
\end{itemize}

\section{Related Work}

\subsection{Self-supervised Video Representation Learning}

There is an increasing interest in learning representation from unlabeled videos.
Many works explore the intrinsic structure of videos and design video-specific pretext tasks, such as estimating video playback rates~\cite{speednet2020,pacepred2020,prp2020}, verifying temporal orders of clips~\cite{shuffle_learn2016,opn2017,vcop2019}, predicting video rotations~\cite{3drotnet2018}, solving space-time cubic puzzles~\cite{stpuzzle2019}, and dense predictive coding~\cite{dpc2019,memorydpc2020}.

Contrastive learning has recently shown great success in image representation learning~\cite{infonce2018,moco2019,simclr2020}, where self-supervised pretraining is approaching the performance of the supervised counterpart~\cite{mocov2_2020,simclrv2_2020}.
More recently, contrastive learning has been introduced to video domain~\cite{cbt2019,vtdl2020,cvrl2020}, where clips from the same video are pulled together while clips from different videos are pushed away.
Another type of contrastive learning methods employ adaptive cluster assignment~\cite{swav2020,sela2020,xdc2019}, where the representation and embedding clusters are simultaneously learned.
However, since most methods apply contrastive learning on raw RGB clips without separating the motion information, the learned representation may suffer from the context bias problem.
This work also follows the contrastive learning paradigm, but we explicitly decouple the motion supervision from the context bias in the pretext task.

Considering the multi-modality of videos, many works explore mutual supervision across modalities to improve the learned representation. For example, they regard the temporal or semantic consistency between videos and the corresponding audios~\cite{atvs2018,xdc2019,avslowfast2020,elo2020}, narrations~\cite{milnce2020}, or a combination of different modalities~\cite{gdt2020} as a natural source of supervision for representation learning.

A recent work named DSM~\cite{dsm2020} shares some similarities with our framework.
It also tries to enhance the learned video representation by decoupling the scene and the motion.
However, it achieves this by simply changing the construction of positive and negative pairs in contrastive learning, and it still learns on raw video clips; while our approach explicitly decouples the context and motion information in the source of supervision.
Besides, the significantly better performance of our approach than DSM on downstream tasks also verifies the superiority of our work.

\subsection{Action Recognition in Compressed Videos}

Video compression techniques (\eg, H.264 and MPEG-4) usually store only a few key frames completely, and reconstruct other frames using motion vectors and residual errors from the key frames.
Taking advantage of this, many works propose to build video models directly on the compressed data to achieve faster inference and better performance~\cite{mv2016,mv2018,coviar2018,dmc2019}.
Pioneering works~\cite{mv2016,mv2018} replace the optical flow stream in two-stream action recognition models~\cite{twostream2014} with a motion vector stream, thereby avoiding slow optical flow extraction.
CoViAR~\cite{coviar2018} further utilizes all modalities (\ie, I-frames, motion vectors, and residuals) in compressed video and bypasses the decoding of RGB frames for efficient video action recognition.
DMC~\cite{dmc2019} improves the quality of the motion vector maps by jointly learning a generative adversarial network.
More recently, Wang \etal~\cite{compressed_detection2019} presents a method for fast object detection in compressed video, showing the potential of using compressed data for more fine-grained tasks.
This work uses MPEG-2 Part2~\cite{video_compression_1991} for video encoding, as practiced in CoViAR, where every I-frame is followed by 11 consecutive P-frames.
We take the I-frames and motion vectors as proxies of the context and motion information for self-supervised video representation learning.

\subsection{Motion Prediction}

Motion prediction task usually refers to deduce the states (\eg, position, orientation, speed, or posture) of an object in a near future~\cite{human_motion_prediction_2015,traj_prediction_2018}. Example applications include human pose prediction~\cite{human_motion_prediction_2015,human_motion_prediction_2017,human_motion_prediction_2019} and traffic prediction~\cite{traj_prediction_2018,traj_prediction_2020}.
Typical models for solving this task include RNNs~\cite{mp_rnn2017,human_motion_prediction_2017}, Transformers~\cite{mp_transformer_2020,mp_transformer_2020_2}, and graph neural networks~\cite{human_motion_prediction_2019}.
In this work, we leverage a simple Transformer encoder-decoder network~\cite{transformer2017} for predicting future motion features based on current visual observations.

\section{Methodology}

\subsection{Overall Framework}
This work presents a self-supervised video representation learning method with two decoupled pretext tasks:
a context matching task for learning coarse-grained context representation;
and a motion prediction task for learning fine-grained motion representation.
For efficiency, we use the I-frames and motion vectors in compressed video as the sources of supervision for context and motion tasks, respectively.
Figure~\ref{fig:overview} shows an overview of our framework, where the V-Network, I-Network, and M-Network are used to extract video, context, and motion features, respectively.

In the context matching task, we compare global features of video clips and I-frames.
An InfoNCE loss~\cite{infonce2018} is cast, where (\textit{video clip, I-frame}) pairs from the same videos are pulled together, while pairs from different videos are pushed away.
By matching still images to video clips, the learned representation is supposed to capture the global and coarse-grained contextual information of the video.

Unlike context, motion information is relatively sparse, localized and fine-grained.
Therefore, we prefer to use a pointwise task to guide the V-Network to capture motion representation.
A simple choice is to estimate motion vectors correspond to a video clip.
However, this would lead the model to learn \textit{low-level} offsets (\eg, optical flow) instead of \textit{high-level} trajectories or actions that are more preferred.
In this work, we use \textit{motion prediction} as the pretext task, where visual data of the current clip is used to predict motion information in a near future.
The assumption is that, to predict motion dynamics in future frames, V-Network needs to extract \textit{low-level} movements from the video and reorganize them into \textit{high-level} trajectories.
In this way, the learned representation can capture long-term motion information, which is beneficial for downstream tasks.

In our implementation, instead of directly predicting the motion vector values, we estimate the features of motion vector maps using an encoder-decoder network (\ie, Transformer~\cite{transformer2017}), and compare the predicted and ``groundtruth'' motion features in a pointwise way.
We find that leads to more stable pretraining and better transferring performance.
The context matching and motion prediction tasks are jointly optimized in an end-to-end fashion.
Next we will introduce these pretext tasks respectively in details.

\subsection{Context Matching}
The context of a video is relatively static in a period of time.
It depicts a global environment in which the action takes place.
We present a \textit{context matching} task for the video model to capture such information, where the source of supervision comes from the I-frames extracted from the compressed video.
A brief overview of the context matching process is shown in the upper half of Figure~\ref{fig:overview}.

Specifically, we extract features $\mathbf{x}_i\in \mathbb{R}^{C_1\times T_1\times H_1\times W_1}$ of a random clip in video $i$,
and features $\mathbf{z}_i\in \mathbb{R}^{C_2\times H_2\times W_2}$ of a random I-frame surrounding the clip.
Then we use global average pooling to obtain their global representation $\mathbf{x}_i'\in \mathbb{R}^{C_1}$ and $\mathbf{z}_i'\in \mathbb{R}^{C_2}$.
Following the design improvements used in recent unsupervised frameworks~\cite{simclr2020,mocov2_2020}, we apply two-layer MLP heads $g^V$ and $g^I$ on the clip and I-frame features respectively to obtain $\mathbf{x}_i^* = g^V(\mathbf{x}_i')\in \mathbb{R}^{C}$ and $\mathbf{z}_i^* = g^I(\mathbf{z}_i')\in \mathbb{R}^{C}$.
Then the InfoNCE loss is applied:
\begin{eqnarray}
J_I = -\frac{1}{B} \sum_{i=1}^B \log\frac{\exp(\cos(\mathbf{z}_i^*, \mathbf{x}_i^*) / \tau)}{\sum_{k=1}^B \exp(\cos(\mathbf{z}_k^*, \mathbf{x}_i^*) / \tau)},
\end{eqnarray}
where $B$ denotes the number of samples in the minibatch, $\cos(\mathbf{z}, \mathbf{x}) = (\mathbf{z}^\mathrm{T} \mathbf{x}) / (\|\mathbf{z}\|_2 \cdot \|\mathbf{x}\|_2)$ denotes the cosine similarity between $\mathbf{z}$ and $\mathbf{x}$, and $\tau$ is a temperature adjusting the scale of cosine similarities.
The loss function pulls video clips and I-frames from the same videos together, and pushes those from different videos far apart.

\subsection{Motion Prediction}
Compared with contextual information, motion information is more fine-grained and position (in $x$, $y$, and $t$ dimensions) sensitive.
To encourage the video network to capture high-level and long-term motion information, we design a motion prediction task where visual data of current clip is used to predict motion dynamics in the near future.
We use motion vectors extracted from the compressed video as the source of supervision.
The lower half of Figure~\ref{fig:overview} shows a brief overview of the motion prediction process.

\begin{table}[t]
\small
\begin{center}
\begin{tabular}{
   >{\raggedright\arraybackslash} m{0.8cm}
   >{\centering\arraybackslash} m{1.35cm}
   >{\centering\arraybackslash} m{2.25cm} |
   >{\centering\arraybackslash} m{0.85cm}
   >{\centering\arraybackslash} m{1.15cm}} \hline
Period & Loss & Encoder-Decoder & UCF101 & HMDB51 \\ \hline
\cellcolor[HTML]{E1E1E1}Current & InfoNCE & Transformer & 29.6 & 10.4 \\
Future & \cellcolor[HTML]{E1E1E1}Cross-Ent. & Transformer & 23.0 & 8.1 \\
Future & \cellcolor[HTML]{E1E1E1}MSE & Transformer & 27.9 & 10.2 \\
Future & InfoNCE & \cellcolor[HTML]{E1E1E1}ConvGRU & 39.8 & 15.2 \\ \hline
Future & InfoNCE & Transformer & \textbf{41.7} & \textbf{16.8} \\ \hline
\end{tabular}
\end{center}
\caption{
   Video retrieval performance (R@$1$) comparison when applying different settings in motion prediction.
   Words with gray background denote the setting changes with respect to our baseline.
   The Mean Square Error (\emph{abbr.} MSE) and Cross Entropy (\emph{abbr.} Cross-Ent.) losses are used when we directly predict motion vector values and value ranges after quantization, respectively.
}
\label{tb:motion_prediction}
\end{table}

Specifically, we extract features $\mathbf{x}_i\in \mathbb{R}^{C_1\times T_1\times H_1\times W_1}$ from a clip of video $i$,
and features $\mathbf{v}_i\in \mathbb{R}^{C_3\times T_3\times H_3\times W_3}$ from the motion vector maps in a near future after the clip.
Subsequently, we feed $\mathbf{x}_i$ to an encoder-decoder network $\mathcal{T}$ (\ie, Transformer~\cite{transformer2017} or ConvGRU~\cite{convgru2015,memorydpc2020}) to predict the motion features $\hat{\mathbf{v}}_i = \mathcal{T}(\mathbf{x}_i)\in \mathbb{R}^{C_3\times T_3\times H_3\times W_3}$.
The ``groundtruth'' and predicted motion features are then flattened and fed into two-layer MLP heads $g_1^M$ and $g_2^M$ respectively to obtain projected embeddings
$\mathbf{v}_i^* = g_1^M(\mathrm{flatten}(\mathbf{v}_i))\in \mathbb{R}^{C\times N}$ and $\hat{\mathbf{v}}_i^* = g_2^M(\mathrm{flatten}(\hat{\mathbf{v}}_i))\in \mathbb{R}^{C\times N}$,
where $N = T_3\cdot H_3\cdot W_3$. The pointwise InfoNCE loss is then applied to $\mathbf{v}_i^*$ and $\hat{\mathbf{v}}_i^*$:
\begin{eqnarray}
\label{eq:pointwise_infonce}
J_M = -\frac{1}{BN} \sum_{i,j} \log\frac{\exp(\cos(\mathbf{v}_{ij}^*, \hat{\mathbf{v}}_{ij}^*) / \tau)}
{\sum_{\substack{k,l}} \exp(\cos(\mathbf{v}_{kl}^*, \hat{\mathbf{v}}_{ij}^*) / \tau)},
\end{eqnarray}
where $i,k=1\sim B,~j,l=1\sim N$, $\mathbf{v}_{ij}^*\in \mathbb{R}^C$ denotes the $j$th column of $\mathbf{v}_i^*$.
In the loss function, only feature points corresponding to the same video $i$ and at the same spatial and temporal position $(x, y, t)$ are regarded as positive pairs, otherwise they are regarded as negative pairs.
The pointwise InfoNCE loss aims to lead the video network to learn fine-grained motion representation.

We test several modeling options and configurations of the motion prediction task, \eg, whether to predict \textit{current} or \textit{future} motion information, whether to match features using the InfoNCE loss or to directly predict motion vector values, and the use of different encoder-decoder networks, \etc
Comparison of these settings is shown in Table~\ref{tb:motion_prediction}.
To summarize, we find that:
1) Predicting \textit{future} motion information leads to significantly better video retrieval performance compared with estimating \textit{current} motion information;
2) Matching predicted and ``groundtruth'' motion features using the pointwise InfoNCE loss brings better results than directly estimating motion vector values;
3) Different encoder-decoder networks lead to similar results, while using Transformer performs slightly better.

In this work, we follow the optimal setting, where we predict future motion features using a Transformer network, and we employ the pointwise InfoNCE loss for training the model.
When applying the Transformer network, we simply consider the input $\mathbf{x}_i\in \mathbb{R}^{C_1\times T_1\times H_1\times W_1}$ as a $1$-D sequence of length $T_1\cdot H_1\cdot W_1$, and the output $\hat{\mathbf{v}}_i\in \mathbb{R}^{C_3\times T_3\times H_3\times W_3}$ as a a $1$-D sequence of length $T_3\cdot H_3\cdot W_3$.

\subsection{Joint Optimization}
We linearly combine the context matching loss and the motion prediction loss to obtain the final loss:
\begin{eqnarray}\label{eq:loss}
J = (1 - \alpha) J_I + \alpha J_M,
\end{eqnarray}
where the $\alpha$ is a scalar hyper-parameter within $[0, 1]$.
We simply set $\alpha = 0.5$ in our experiments, where the context and motion losses are equally weighted.
The V-Network, I-Network, M-Network, Transformer $\mathcal{T}$, and all MLP heads (\ie, $g^V$, $g^I$, $g^M_1$, and $g^M_2$) are jointly optimized with loss function \eqref{eq:loss} in an end-to-end fashion.

\section{Experiments}

\subsection{Experiment Settings}

\noindent\textbf{Datasets.}
All experiments in this paper are conducted on three video classification datasets: UCF101~\cite{ucf101}, HMDB51~\cite{hmdb51}, and Kinetics400~\cite{kinetics400}.
UCF101 consists of 13,320 videos belonging to 101 classes, while HMDB51 consists of 6,766 videos in 51 classes.
Both datasets are divided into three train/test splits.
We use their first split in all our experiments.
Kinetics400 is a large-scale dataset containing 246K/20K videos in its train/val subsets.
It populates 400 classes of human actions.

\noindent\textbf{Networks.}
We evaluate the performance of our framework based on three video backbones (also the V-Network in Figure~\ref{fig:overview}): C3D~\cite{c3d2015}, R(2+1)D-26~\cite{resnet2p1d_2018}, and R3D-26~\cite{resnet2016,resnet2p1d_2018}.
For the I-Network and M-Network, we use shallow R2D-10 and R3D-10 backbones, respectively, where each of them comprises $4$ layers of ResNet BasicBlocks~\cite{resnet2016,resnet2p1d_2018}.
The encoder-decoder network in our framework is implemented as a shallow Transformer, where 2 encoding layers and 4 decoding layers are used.
Feature dimensions in the hidden layers of MLP heads are set to $2048$, while dimensions in the hidden layers of the Transformer network and the output layers of MLP heads are set to $512$.

\setlength{\tabcolsep}{4pt}
\begin{table}[t]
\small
\centering
\begin{tabular}{l|ccc|ccc}
\hline
\multirow{2}{*}{Backbone} & \multicolumn{3}{c|}{UCF101} & \multicolumn{3}{c}{HMDB51} \\ \cline{2-7}
& Scratch & UCF & K400 & Scratch & UCF & K400 \\
\hline
C3D & 60.5 & 78.6 & \textbf{83.4} & 29.2 & 46.9 & \textbf{52.9} \\
R(2+1)D-26 & 65.0 & 79.7 & \textbf{85.7} & 32.5 & 48.6 & \textbf{54.0} \\
R3D-26 & 58.0 & 76.6 & \textbf{83.7} & 28.9 & 47.2 & \textbf{55.2} \\
\hline
\end{tabular}
\vspace{5pt}
\caption{
   Action recognition performance (\ie, top-1 accuracy) comparison between training from scratch and from our pretrained models.
   The three video backbones are unsupervised pretrained on either UCF101 (\emph{abbr.} UCF) or Kinetics400 (\emph{abbr.} K400) datasets.
   }
\label{tab:scratch_pretrain}
\end{table}

\setlength{\tabcolsep}{6pt}
\begin{table}[t]
\small
\centering
\begin{tabular}{l|cccccc} \hline
\multirow{2}{*}{Pretraining} & \multicolumn{6}{c}{Finetuning epoch} \\ \cline{2-7}
& 1 & 5 & 10 & 20 & 50 & 120 \\ \hline
Scratch & 3.2 & 17.5 & 22.4 & 28.9 & 42.7 & 65.0 \\
UCF101 & 6.1 & 17.0 & 43.2 & 52.7 & 64.3 & 79.7 \\
Kinetics400 & \textbf{11.5} & \textbf{41.8} & \textbf{54.4} & \textbf{65.1} & \textbf{73.2} & \textbf{85.7} \\ \hline
\end{tabular}
\vspace{5pt}
\caption{
   Convergence speed comparison between training from scratch and from pretrained models.
   The top-1 accuracy of the R(2+1)D-26 network on the UCF101 action recognition task is used as the indicator.
}
\label{tab:convergence}
\end{table}

\noindent\textbf{Pretraining settings.}
Two datasets are used for pretraining: either UCF101 or Kinetics400.
For UCF101, we pretrain our framework for 120 epochs on 4 GPUs, with a total batch size of 64.
For Kinetics400, we pretrain our framework for 120 epochs on 32 GPUs, with a total batch size of 512.
Following recent practices in large-batch training~\cite{linear_lr_2017}, we set the learning rate ($\mathrm{lr}$) to scale up linearly with the batch size: $\mathrm{lr} = 0.0005\times B$.
A cosine annealing rule is applied to decay the learning rate smoothly during training.
We use SGD as the optimizer, where the weight decay and momentum are set to $0.005$ and $0.9$, respectively.
Each video clip consists of 16 frames with a temporal stride of 4, and we predict motion dynamics in the next 8 consecutive frames.
All clips are resized to $16\times C_{\mathrm{in}}\times 112\times 112$, where $C_{\mathrm{in}} = 3$ for video clips and $C_{\mathrm{in}} = 2$ for motion vector maps.
To introduce hard negatives for each video clip, we sample one positive clip and three negative clips of motion vector maps from the same video.
We use random crop, random flip, Gaussian blur, and color jitter as the data augmentation for video clips and I-frames, and we use random crop and random flip as the data augmentation for motion vector maps.
We ensure that the cropping and flipping parameters are consistent for positive (\textit{video clip}, \textit{motion vectors}) pairs.

\noindent\textbf{Finetuning settings.}
Pretrained models are finetuned on either UCF101 or HMDB51 to assess the transferring performance.
For UCF101, we set the learning rate as $\mathrm{lr} = 0.0001\times B$ and the weight decay of the SGD optimizer as $0.003$.
For HMDB51, we set the learning rate as $\mathrm{lr} = 0.0002\times B$ and the weight decay as $0.002$.
For both datasets, we finetune the model for 120 epochs on 1 GPU with a batch size of $8$.
We use the cosine annealing scheduler to decay $\mathrm{lr}$ during training.
A dropout of $0.3$ is used before feeding backbone features to the classifier.
We use random crop, random flip, Gaussian blur, and color jitter as the augmentation to improve the diversity of training data.
As a common practice, we uniformly sample 10 clips in a video and average their scores for performance evaluation.

\noindent\textbf{Video retrieval settings.}
Video retrieval experiments are conducted on either UCF101 or HMDB51 datasets, where videos in the test set are used to find videos in the training set.
The \textit{recall at top-$k$} (\emph{abbr.} R@$k$) is used as the metric for evaluation -- if one of the top-$k$ searched videos has the same class label as the query video, a successful retrieval is count.
Following the practice of~\cite{vcop2019,memorydpc2020}, we sample 10 video clips with a sliding window, and use the average of their global features as the representation of the video.

\subsection{Ablation Study}

\setlength{\tabcolsep}{8pt}
\begin{table}[t]
\small
\centering
\begin{tabular}{l|cc|cc} \hline
\multirow{2}{*}{Modality} & \multicolumn{2}{c|}{UCF101} & \multicolumn{2}{c}{HMDB51} \\ \cline{2-5}
& R@$1$ & R@$5$ & R@$1$ & R@$5$ \\ \hline
I-frames & 33.8 & 47.7 & 11.5 & 28.3 \\
Motion vectors  & 30.4 & 50.9 & 14.0 & 37.5 \\
I + Optical flow & \textbf{43.2} & \textbf{58.9} & \textbf{17.7} & \textbf{38.5} \\ \hline
I + Motion vectors & \textbf{41.7} & \textbf{57.4} & \textbf{16.8} & \textbf{37.2} \\ \hline
\end{tabular}
\vspace{5pt}
\caption{
   Video retrieval performance comparison when pretraining C3D network on the UCF101 dataset using different modalities as the source of supervision.
   We also evaluate the results using optical flow here for a reference.
}
\label{tab:modality}
\end{table}

\setlength{\tabcolsep}{11pt}
\begin{table}[t]
\small
\centering
\begin{tabular}{l|cc|cc} \hline
\multirow{2}{*}{Network} & \multicolumn{2}{c|}{UCF101} & \multicolumn{2}{c}{HMDB51} \\ \cline{2-5}
& R@$1$ & R@$5$ & R@$1$ & R@$5$ \\ \hline
I-Network & \textbf{32.3} & \textbf{47.9} & \textbf{12.0} & 27.8 \\
M-Network & 25.4 & 44.2 & 10.7 & \textbf{32.1} \\ \hline
\textcolor{gray}{V-Network} & \textcolor{gray}{41.7} & \textcolor{gray}{57.4} & \textcolor{gray}{16.8} & \textcolor{gray}{37.2} \\ \hline
\end{tabular}
\vspace{5pt}
\caption{
   Video retrieval performance comparison when using the pretrained I-Network and M-Network for the retrieval.
   We list the results of V-Network as a reference.
   The pretraining experiment is conducted on the UCF101 dataset with C3D as the video backbone.
}
\label{tab:i_m_networks}
\end{table}

This section validates several modeling and configuration options in our framework.
The finetuning or video retrieval results on UCF101 and HMDB51 datasets are used as the performance indicators.

\noindent\textbf{Ablation: scratch \emph{vs.} pretraining.}
Table~\ref{tab:scratch_pretrain} compares the action recognition performance of three different backbones on UCF101 and HMDB51 datasets when training from scratch (\ie, from randomly initialized parameters) and from our pretrained models.
We observe that:
1) Even without introducing new training data, self-supervised pretraining followed by supervised finetuning on UCF101 leads to a remarkable $14.7\%\sim 18.6\%$ performance gain for all three backbones, suggesting that pretraining with context and motion decoupling significantly improves the quality of the learned representation;
2) Training on larger Kinetics400 ($\sim 25\times$ the scale of UCF101) further improves the accuracy with a margin of $4.8\%\sim 7.1\%$ on UCF101 and a margin of $5.4\%\sim 8.0\%$ on HMDB51, showing the scalability of our framework on larger-scale datasets.
We also compare the convergence speed when training from scratch and from pretrained models on UCF101.
Results are shown in Table~\ref{tab:convergence}.
We observe that self-supervised pretraining clearly boosts the convergence, especially when pretrained on the larger Kinetics400 dataset, where the top-1 accuracy on epoch $5$ surpasses that without pretraining by $24.3\%$.

\setlength{\tabcolsep}{7pt}
\begin{table}[t]
\small
\centering
\begin{tabular}{l|cc|cc} \hline
\multirow{2}{*}{Method} & \multicolumn{2}{c}{UCF101} & \multicolumn{2}{c}{HMDB51} \\ \cline{2-5}
& Top1 & Top5 & Top1 & Top5 \\ \hline
C3D & 60.5 & 84.2 & 29.2 & 58.0 \\
C3D with Reg. & \textbf{74.3} & \textbf{91.4} & \textbf{38.5} & \textbf{69.0} \\ \hline
R(2+1)D-26 & 65.0 & 85.9 & 32.5 & 66.1 \\
R(2+1)D-26 with Reg. & \textbf{75.6} & \textbf{92.3} & \textbf{41.7} & \textbf{71.3} \\ \hline
R3D-26 & 58.0 & 83.1 & 28.9 & 60.4 \\
R3D-26 with Reg. & \textbf{71.2} & \textbf{88.9} & \textbf{36.3} & \textbf{67.9} \\ \hline
\end{tabular}
\vspace{5pt}
\caption{
   Action recognition performance comparison of three backbones trained with and without the auxiliary motion prediction task as regularization.
}
\label{tab:regularization}
\end{table}

\setlength{\tabcolsep}{5pt}
\begin{table}[t]
\small
\centering
\begin{tabular}{l|cccccc} \hline
\multirow{2}{*}{Task} & \multicolumn{6}{c}{Pretraining epoch} \\ \cline{2-7}
& 1 & 5 & 10 & 20 & 50 & 120 \\ \hline
Context matching & 25.8 & 42.5 & 48.0 & 50.0 & 53.7 & \textbf{65.9} \\
Motion prediction & 0.1 & 1.5 & 4.1 & 8.4 & 15.9 & \textbf{33.1} \\ \hline
Video retrieval & 6.0 & 6.3 & 8.5 & 9.5 & 14.8 & \textbf{16.8} \\ \hline
\end{tabular}
\vspace{5pt}
\caption{
   Correlation between pretraining and transferring tasks.
   The top-1 accuracy of the context matching and motion prediction tasks, as well as the R@1 score of the downstream video retrieval task are recorded during the pretraining process.
   The pretraining experiment is conducted on the UCF101 dataset with C3D as the video backbones, while the evaluation is performed on the HMDB51 dataset.
}
\label{tab:correlation}
\end{table}

\setlength{\tabcolsep}{8pt}
\begin{table*}[!htb]
\small
\centering
\begin{tabu}{ll|lcc|cc} \hline
Method & Year & Pretrained & Resolution & Architecture & UCF101 & HMDB51 \\ \hline
Shuffle \& Learn~\cite{shuffle_learn2016} & 2016\hspace{1pt} & UCF101 & $227\times 227$ & CaffeNet & 50.2 & 18.1 \\
OPN~\cite{opn2017} & 2017\hspace{1pt} & UCF101 & $227\times 227$ & VGG-14 & 59.6 & 23.8 \\
DPC~\cite{dpc2019} & 2019\hspace{1pt} & UCF101 & $128\times 128$ & R3D-18 & 60.6 & - \\
VCOP~\cite{vcop2019} & 2019\hspace{1pt} & UCF101 & $112\times 112$ & R(2+1)D-26 & 72.4 & 30.9 \\
PacePred~\cite{pacepred2020} & 2020\hspace{1pt} & UCF101 & $112\times 112$ & R(2+1)D-18 & \textbf{75.9} & 35.9 \\
VTDL~\cite{vtdl2020} & 2020\hspace{1pt} & UCF101 & $112\times 112$ & C3D & 73.2 & \textbf{40.6} \\
PRP~\cite{prp2020} & 2020\hspace{1pt} & UCF101 & $112\times 112$ & C3D & 69.1 & 34.5 \\
VCP~\cite{vcp2020} & 2020\hspace{1pt} & UCF101 & $112\times 112$ & C3D & 68.5 & 32.5 \\
DSM~\cite{dsm2020} & 2020\hspace{1pt} & UCF101 & $112\times 112$ & C3D & 70.3 & 40.5 \\
\hline
\textbf{Ours} & \hspace{1pt} & UCF101 & $112\times 112$ & \textbf{C3D} & \textbf{78.6} & \textbf{46.9} \\
\textbf{Ours} & \hspace{1pt} & UCF101 & $112\times 112$ & \textbf{R(2+1)D-26} & \textbf{79.7} & \textbf{48.6} \\
\textbf{Ours} & \hspace{1pt} & UCF101 & $112\times 112$ & \textbf{R3D-26} & \textbf{76.6} & \textbf{47.2} \\
\hline \hline
3D-RotNet~\cite{3drotnet2018} & 2018\hspace{1pt} & Kinetics400 & $112\times 112$ & R3D-18 & 62.9 & 33.7 \\
ST-Puzzle~\cite{stpuzzle2019} & 2019\hspace{1pt} & Kinetics400 & $224\times 224$ & R3D-18 & 65.8 & 33.7 \\
DPC~\cite{dpc2019} & 2019\hspace{1pt} & Kinetics400 & $128\times 128$ & R3D-18 & 68.2 & 34.5 \\
DPC~\cite{dpc2019} & 2019\hspace{1pt} & Kinetics400 & $224\times 224$ & R3D-34 & 75.7 & 35.7 \\
CBT~\cite{cbt2019} & 2019\hspace{1pt} & Kinetics600 & $112\times 112$ & S3D-G & 79.5 & 44.6 \\
SpeedNet~\cite{speednet2020} & 2020\hspace{1pt} & Kinetics400 & $224\times 224$ & S3D-G & 81.1 & 48.8 \\
MemoryDPC~\cite{memorydpc2020} & 2020\hspace{1pt} & Kinetics400 & $224\times 224$ & R2D3D-34 & 78.1 & 41.2 \\
PacePred~\cite{pacepred2020} & 2020\hspace{1pt} & Kinetics400 & $112\times 112$ & R(2+1)D-18 & 77.1 & 36.6 \\
VTDL~\cite{vtdl2020} & 2020\hspace{1pt} & Kinetics400 & $112\times 112$ & C3D & 75.5 & 43.2 \\
DSM~\cite{dsm2020} & 2020\hspace{1pt} & Kinetics400 & $224\times 224$ & I3D & 74.8 & 52.5 \\
DSM~\cite{dsm2020} & 2020\hspace{1pt} & Kinetics400 & $224\times 224$ & R3D-34 & 78.2 & \textbf{52.8} \\
VTHCL~\cite{vthcl2020} & 2020\hspace{1pt} & Kinetics400 & - & R3D-50 & \textbf{82.1} & 49.2 \\
\hline
\textbf{Ours} & \hspace{1pt} & Kinetics400 & $112\times 112$ & \textbf{C3D} & \textbf{83.4} & \textbf{52.9} \\
\textbf{Ours} & \hspace{1pt} & Kinetics400 & $112\times 112$ & \textbf{R(2+1)D-26} & \textbf{85.7} & \textbf{54.0} \\
\textbf{Ours} & \hspace{1pt} & Kinetics400 & $112\times 112$ & \textbf{R3D-26} & \textbf{83.7} & \textbf{55.2} \\
\hline
\end{tabu}
\vspace{5pt}
\caption{Comparison with state-of-the-art self-supervised approaches on action recognition on UCF101 and HMDB51.}
\label{tab:sota_recognition}
\end{table*}

\setlength{\tabcolsep}{7pt}
\begin{table*}[h!]
\small
\centering
\begin{tabular}{lcc|cccc|cccc}
\hline
\multirow{2}{*}{Method} & \multirow{2}{*}{Year} & \multirow{2}{*}{Pretrained} & \multicolumn{4}{c|}{UCF101} & \multicolumn{4}{c}{HMDB} \\ \cline{4-11}
& & & R@1 & R@5 & R@10 & R@20  & R@1 & R@5 & R@10 & R@20 \\ \hline
Jigsaw~\cite{jigsaw2016} & 2016 & UCF101 & 19.7 & 28.5 & 33.5  & 40.0 & - & - & - & - \\
OPN~\cite{opn2017} & 2017 & UCF101 & 19.9 & 28.7 & 34.0 & 40.6 & - & - & - & - \\
Buchler~\cite{buchler2018} & 2018 & UCF101 & \textbf{25.7} & 36.2 & 42.2  & 49.2 & - & - & - & - \\
VCOP~\cite{vcop2019} & 2019 & UCF101 & 14.1 & 30.3 & 40.4  & 51.1 & 7.6 & 22.9 & 34.4 & 48.8 \\
VCP~\cite{vcp2020} & 2020 & UCF101 & 18.6 & 33.6 & 42.5 & 53.5 & 7.6 & 24.4 & 36.3 & 53.6 \\
MemoryDPC~\cite{memorydpc2020} & 2020 & UCF101 & 20.2 & \textbf{40.4} & \textbf{52.4} & \textbf{64.7} & \textbf{7.7} & \textbf{25.7} & \textbf{40.6} & \textbf{57.7} \\
SpeedNet~\cite{speednet2020} & 2020 & Kinetics400 & 13.0 & 28.1 & 37.5 & 49.5 & - & - & - & - \\
\hline
\textbf{Ours (C3D)} & & UCF101 & \textbf{41.7} & \textbf{57.4} & \textbf{66.9} & \textbf{76.1} & \textbf{16.8} & \textbf{37.2} & 50.0 & 64.3 \\
\textbf{Ours (R(2+1)D-26)} & & UCF101 & 38.4 & 55.4 & 65.2 & 74.6 & 14.3 & 36.0 & 48.5 & 64.2 \\
\textbf{Ours (R3D-26)} & & UCF101 & 38.9 & 56.1 & 65.8 & 75.6 & 15.2 & 36.0 & \textbf{51.4} & \textbf{65.5} \\
\hline
\end{tabular}
\vspace{5pt}
\caption{Comparison with state-of-the-art self-supervised approaches on video retrieval on UCF101 and HMDB51.}
\label{tab:sota_retrieval}
\end{table*}

\noindent\textbf{Ablation: pretraining with different modalities.}
Table~\ref{tab:modality} compares the video retrieval performance when pretraining C3D network on UCF101 using different modalities as the source of supervision.
As a reference, we also introduce the optical flow~\cite{tvl1_2007} as a substitute for the motion vector maps, which is more accurate but also more computationally expensive to extract.
We find that:
1) Context-only pretraining brings higher R@1 score on UCF101, while motion-only pretraining leads to better results on HMDB51.
This is consistent with the observation that most classes in UCF101 can be distinguished using static context or pose cues, while classes in HMDB51 mainly differ in motion~\cite{hmdb51};
2) Learning both context and motion information leads to much better results than learning any of them alone, suggesting the complementarity between the two modalities;
3) Replacing motion vectors with optical flow brings only slight performance gains, but motion vectors are several orders of magnitudes faster to extract.

\noindent\textbf{Ablation: effectiveness of I-Network and M-Network.}
To assess if the co-trained I-Network and M-Network also learned effective representation, we evaluate the video retrieval performance on UCF101 and HMDB51 using the pretrained I-Network or M-Network.
The pretraining experiment is conducted with C3D as the video backbone.
Results are shown in Table~\ref{tab:i_m_networks}.
We observe that both networks can obtain reasonable video retrieval recalls.
Interestingly, we find that I-Network performs much better than M-Network on UCF101 with a gain of $6.9\%$, and is comparable with M-Network on HMDB51 (with a gain of $1.3\%$ in R@$1$ and a reduction of $4.3\%$ in R@5).
This may be because most classes in UCF101 can be classified by context information, while action classes in HMDB51 are mainly distinguished by the motion information~\cite{hmdb51}.

\noindent\textbf{Ablation: motion prediction as regularization.}
Spatial-temporal neural networks often learn context bias rather than motion dynamics~\cite{diving48_2018}, leading to worse results on action classes that heavily depend on motion patterns.
We assess whether explicitly introducing an auxiliary motion prediction task improves the performance.
Results are shown in Table~\ref{tab:regularization}.
We train three video backbones from scratch on UCF101 and HMDB51, with or without introducing the motion prediction task.
The weight of the pointwise InfoNCE loss (\ie, Eq.~\eqref{eq:pointwise_infonce}) is set to $0.2$.
We observe in Table~\ref{tab:regularization} that introducing motion prediction as regularization consistently improves the action recognition performance,
where the results on UCF101 are improved by $13.8\%, 10.6\%$, and $13.2\%$ respectively for C3D, R(2+1)D-26, and R3D-26 backbones,
and the results on HMDB51 are improved by $9.3\%, 9.2\%$, and $7.4\%$ respectively.

\noindent\textbf{Ablation: correlation between pretraining and transferring.}
To assess whether the pretraining and transferring tasks have high correlation, we record the top-1 accuracies of context matching, motion prediction, and the R@$1$ scores of video retrieval during the pretraining process.
Results are shown in Table~\ref{tab:correlation}.
In the experiment, we pretrain the C3D backbone using our method on UCF101, and evaluate the video retrieval recalls on HMDB51.
We observe that, as the accuracies of the pretraining tasks increase, the transferring performance consistently improves.
This validates the effectiveness of our pretext tasks in self-supervised video representation learning.

\subsection{Comparison with State-of-the-art Approaches}

\noindent\textbf{Transfer learning.}
Table~\ref{tab:sota_recognition} compares our work with previous self-supervised video representation learning methods on UCF101 and HMDB51.
The models are pretrained on either UCF101 or Kinetics400.
To make the comparison as fair as possible, we test our framework with three different backbones (\ie, C3D, R(2+1)D-26, and R3D-26), and we use a unified clip size of $112\times 112$.
As shown in the table, when pretrained on UCF101, our approach significantly outperforms state-of-the-art methods under all three backbones.
Compared with previous best results, we achieve improvements of $2.7\%, 3.8\%$, and $0.7\%$ respectively for the three backbones on UCF101,
and we achieve improvements of $6.3\%, 8.0\%$, and $6.6\%$ respectively on HMDB51.
When pretrained on Kinetics400, we also observe absolute gains of $1.3\%, 3.6\%$, and $1.6\%$ for the three backbones on UCF101, and gains of $0.1\%, 1.2\%$, and $2.4\%$ on HMDB51 over previous methods.
Besides, our models pretrained on Kinetics400 clearly outperforms those pretrained on UCF101, with nonnegligible improvements of $4.8\%\sim 7.1\%$ on UCF101 and $5.4\%\sim 8.0\%$ on HMDB51, indicating the scalability of our approach on larger datasets.

\noindent\textbf{Video retrieval.}
Table~\ref{tab:sota_retrieval} compares the video retrieval recalls of our work and state-of-the-art methods on UCF101 and HMDB51.
All methods except SpeedNet~\cite{speednet2020} are pretrained on the UCF101 dataset, while SpeedNet is pretrained on the Kinetics400 dataset.
Following~\cite{vcop2019,memorydpc2020}, we use videos in the test set as queries to search videos in the training set.
The recall at top-$k$ (R@$k$) as used as the performance indicator.
As shown in the table, the results of our approach outperform previous methods by a very large margin.
Specifically, on UCF101, our R@1 scores surpass state-of-the-art results by $16.0\%, 12.7\%$, and $13.2\%$ when using C3D, R(2+1)D-26, and R3D-26 backbones, respectively;
while on HMDB51, we outperform state-of-the-art results by $11.1\%, 6.7\%$, and $8.5\%$ with the three backbones, respectively.
The results validate the high quality of the video representation learned by our context-motion decoupled self-supervised learning framework.

\section{Conclusion and Future Work}
This paper presents a self-supervised video representation learning framework that explicitly decouples the context and motion supervision in the pretext task.
We take the I-frames and motion vectors in compressed videos as the supervision sources for context and motion, respectively, which can be effectively extracted at more than 500 fps on CPU.
We then present two pretext tasks that are jointly optimized:
a \textit{context matching} task that compares global features of I-frames and video clips under the contrastive learning framework;
and a \textit{motion prediction} task where current visual data in a video are used to predict the future motion information.
The former aims to lead the video network to learn coarse-grained contextual representation, while the latter encourages the video network to capture fine-grained motion representation.
Extensive experiments show that our decoupled learning framework achieves significantly better performance on downstream tasks of both action recognition and video retrieval.
Various ablation studies also verify our several modeling and configuration options.

In the future, we would like to explore how the \textit{residual errors} in compressed video can be used to improve the learned representation.
Residual errors are supplementary information of motion vectors, which indicate the vanishing and emerging pixels and compensate for the estimation errors of motion vectors.
This information may potentially improve the quality of the learned video representation.




{\small
\bibliographystyle{ieee_fullname}
\bibliography{cvpr21_compression_ssl}
}

\end{document}